\documentclass[11pt]{article}

\usepackage[final]{acl}

\usepackage{times}
\usepackage{latexsym}

\usepackage[T1]{fontenc}

\usepackage[utf8]{inputenc}

\usepackage{microtype}

\usepackage{inconsolata}

\usepackage{graphicx}

\usepackage{amsmath}
\usepackage{amssymb}
\usepackage{amsthm}

\title{Informational Antilocality and the Locality Bias in LLMs}

\author{Andrew McInnerney$^{1}$\thanks{\hspace{5pt}Authors contributed equally to this work.} \hspace{15pt} Shane Storks$^{2}$\footnotemark[1]  \hspace{15pt} Steven Abney$^1$ \hspace{15pt} Richard L. Lewis$^1$ \\
       $^1$University of Michigan \hspace{20pt} $^2$Eastern Michigan University \\
        \texttt{\{amcin,abney,rickl\}@umich.edu} \hspace{20pt} \texttt{sstorks@emich.edu} \\ }

\begin{document}
\maketitle
\begin{abstract}
We consider the ability of transformer-based language models (LLMs) to learn what we call $k$\textit{-antilocal} languages, i.e., languages that have no mutual information across any span of $k$ contiguous symbols. We construct such languages with increasing $k$, finding that LLMs trained on them achieve comparable cross-entropy loss regardless of antilocality, but converge more slowly on more antilocal languages. Our findings support the idea that non-local dependencies are more difficult to learn, but the evidence for this bias comes from learning speed rather than learning success.
\end{abstract}

\section{Introduction}

In recent years, Large Language Models (LLMs) have attracted significant scientific interest due to their ability to use natural language in remarkably humanlike ways. Prominent in the cognitive science of LLMs is the question to what extent the mechanisms underlying this linguistic behavior are similar (at the appropriate level of analysis) to those of humans. The literature attests a range of positions in this area, with some authors claiming that linguistic cognition in humans and LLMs are completely disanalogous \citep[e.g.,][]{Chomskyetal:2023}, while others contend that LLMs represent a major step forward in modeling linguistic cognition \citep[e.g.,][]{SartoriOrru:2023, Piantadosi:2024}.

An important argument in this discussion concerns humans' and LLMs' \textit{limitations} with respect to the classes of language they can learn. In theoretical linguistics (particularly in the generative tradition following \citealt{Chomsky:1965}), it has long been assumed that human language acquisition is constrained by a set of innate learning biases (``Universal Grammar'') such that languages with certain unnatural properties are harder or impossible for humans to learn via developmentally typical mechanisms \citep[e.g.,][]{Lenneberg:1967, Chomsky:2000, Moro:2016, FriedericiEtAl:2017}. Much skepticism towards LLMs as models of human linguistic cognition centers around the supposition that LLMs are subject to radically different linguistic biases than humans; it is argued that such differences undermine claims of deep mechanistic analogies between these two linguistic systems \citep{Chomskyetal:2023, MoroEtAl:2023, Bowers:2026}.

\begin{figure}[t]

  \includegraphics[width=0.92\columnwidth]{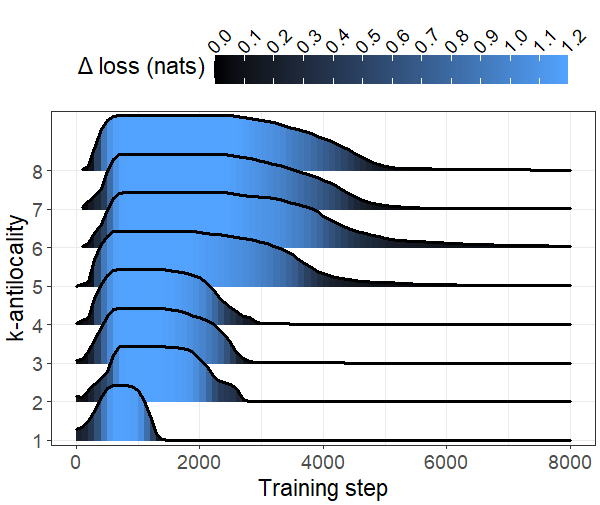}

  \vspace{-5pt}
  
  \caption{Difference in GPT-2 validation cross-entropy loss while training on antilocal vs. control (i.e., simple trigram) languages (antilocal $-$ control). Languages with higher values of $k$ are more antilocal (see Section \ref{sec:antilocal_languages}), hence exhibit less information locality than the corresponding control languages. GPT-2 eventually achieves optimal performance on all languages.}
  \label{fig:difference_plot}
\end{figure}

Studies have found that LLMs can learn ``impossible languages'' of different kinds as easily as they learn natural languages \citep{YangEtAl:2025, ZivEtAl:2026, BowersMitchell:2026}, where the metric for learning success is typically based on cross-entropy loss or KL divergence. However, one constraint that LLMs and humans have been argued to share is \textit{information locality}: languages with less local statistical structure are harder to learn \citep{KalliniEtAl:2024, SomeyaEtAl:2025}.

This bias has a natural source in humans (i.e., human memory degrades over time; see, e.g., \citealt[p52]{Futrell:2017}), but we find it surprising in LLMs. Unlike humans, transformers enjoy simultaneous access to all tokens in their input. Furthermore, attention is entirely parallel; linear distance between tokens (i.e., relative locality) must be learned from artificially injected positional encodings. The source of the locality bias in LLMs is therefore not obvious. The possibility remains that the findings of \citeauthor{KalliniEtAl:2024} and \citeauthor{SomeyaEtAl:2025} can be explained by inductive biases in LLMs that \textit{correlate} with information locality but are ultimately \textit{dissociable} from it \citep[compare][p28003]{SomeyaEtAl:2025}.

In this paper, we define a discrete measure of information (anti-)locality, construct a set of simple languages with variable levels of antilocality, and train LLMs on them. We find that our notion of antilocality does not limit achieved cross-entropy loss for the simple languages we consider, though it does have a marked effect on speed of convergence. We conclude by speculating that LLMs' locality bias may depend specifically on autoregression, because it appears to disappear in a masked LLM.
\section{Related Work}


\citet{KalliniEtAl:2024} found that, compared to a corpus of attested English, transformer-based LLMs struggle to learn various perturbed forms (e.g., sentence-level shuffles or reversals) in terms of both learning success (i.e., minimum achieved perplexity) and learning speed. 
This implies that LLMs, despite being largely data-driven, instantiate at least some relevant human language-learning biases. Because this conclusion potentially challenges the argument from generative grammarians against LLMs-as-cognitive-models outlined above, there has recently been a flurry of interest in LLMs' ability to learn such \textit{impossible languages}. 

Subsequent work has presented mixed results. 
\citet{ZivEtAl:2026} tested additional natural languages, 
finding that in most cases, LLMs could \textit{not} distinguish attested and impossible languages, nor could variance in ease-of-learning metrics be used to cluster them. Using parallel multilingual corpora, 
\citet{YangEtAl:2025} found that while LLMs could largely distinguish possible and impossible forms of the same languages, some attested forms were harder to learn than some impossible forms across languages. They also found that LLMs could not distinguish attested and typologically implausible word orders, suggesting that while LLMs may share some learning biases with humans, this does not systematically apply to those that shape natural linguistic typology.
Focusing on word-order harmony rather than random permutations, \citet{XuEtAl:2026} found that 
while LLMs reached similar perplexity on attested and perturbed languages, perturbed languages were learned more slowly, suggesting LLMs 
are still somewhat typologically aligned, albeit not globally.


In these works, the most difficult languages manipulate the local statistical structure of languages (e.g., deterministic shuffles), suggesting information locality as a learning bias for LLMs.
\citet{BowersMitchell:2026}, however, argue that such randomly perturbed languages lack the sort of structure that natural language has.
\citet{SomeyaEtAl:2025} explore this further, confirming that, with global entropy held constant, LLMs are indeed less successful at learning languages with higher ``local'' entropy (based on next-symbol uncertainty given a fixed context window).
We extend this inquiry by developing synthetic \textit{antilocal} languages, such that tokens that precede the current position by fewer than $k$ tokens provide no information about the next token, enabling precise study of locality effects.
\section{Antilocal Languages}\label{sec:antilocal_languages}

We explore a discrete measure of information locality that we call $k$-antilocality, defined as follows. Consider a language $\mathcal{L}$ formed on vocabulary $V$ (we take $\mathcal{L}$ to be a probability distribution on $V^*$). Let $W$ be a random variable over symbols $w\in V$, and let $C_n$ be a random variable over $n$-length contexts for $W$. Then $\mathcal{L}$ is $k$-antilocal iff
\begin{equation}\label{k_antilocality}
    I(W;C_n)=0\text{ for all }n \leq k
\end{equation}

\noindent The intuition is that $k$-antilocality places a lower-bound of $k$ on dependency length in a language. Another way to put it is that, in a $k$-antilocal language, a context of $k$ or fewer contiguous symbols can never meaningfully contribute to the prediction of an adjacent symbol. For the purposes of this paper, we consider languages whose $n$-gram distributions are uniform for all $n\leq k+1$. We show that such languages are $k$-antilocal in Appendix~\ref{apx:antilocality}.

\begin{figure}[t]
  \includegraphics[width=\columnwidth]{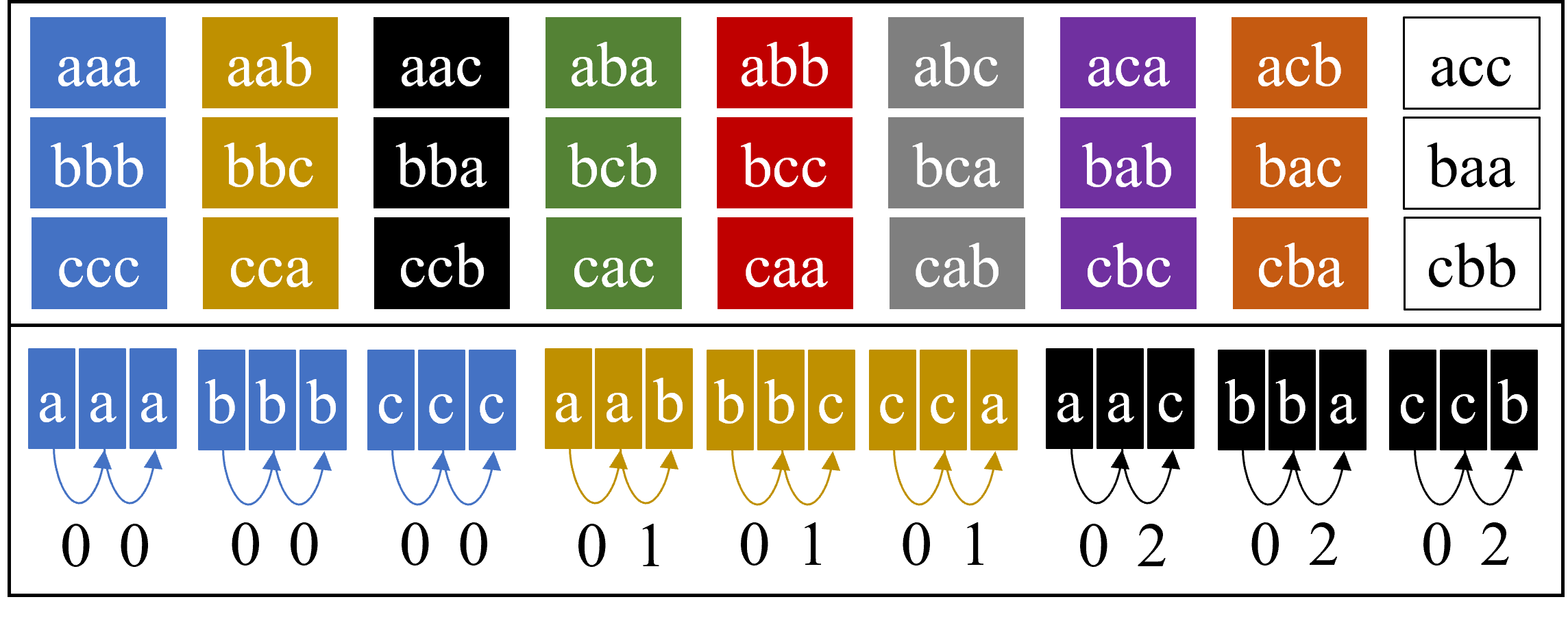}

  \vspace{-5pt}
  
  \caption{Rotation classes from trigrams on vocabulary $V=\{a,b,c\}$. \textbf{Top:} all $|V|^3=27$ trigrams partitioned into rotation classes by color. Note that there are $|V|^{3-1}=9$ rotation classes in total. \textbf{Bottom:} an illustration of symbol-transition distance within classes.}
  \label{fig:rotation_classes}
\end{figure}

We construct eight pairs of languages with this property, described below, which we call $k$-back languages, all derived from simple trigram languages. In each case, given an ordered vocabulary $V$, we start by partitioning the set of trigrams $V^3$ into \textit{rotation classes} defined by rotational/cyclic distance of symbol transitions (i.e., a transition from the $i^{\text{th}}$ vocab item to the $j^{\text{th}}$ has distance $(i-j)\bmod|V|$). This notion is illustrated in Figure \ref{fig:rotation_classes}. To construct our languages, we select a subset of rotation classes to serve as trigram generators. Note that, by partitioning according to rotation class, we ensure uniform unigram distribution within and across generators (see Appendix~\ref{apx:antilocality}).

\begin{figure}[t]

    \centering

  \includegraphics[width=0.92\columnwidth]{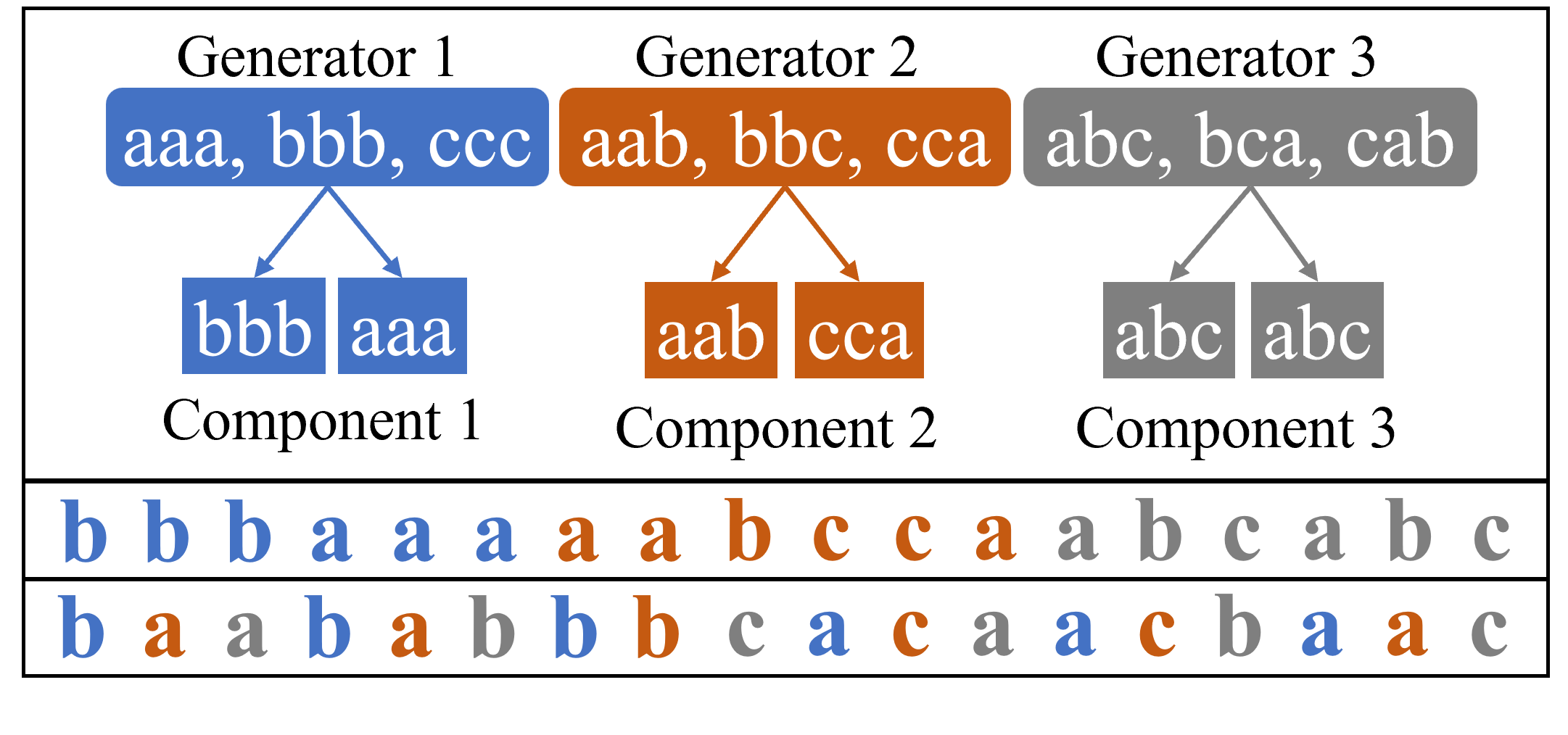}

  \vspace{-7pt}
  
  \caption{Sampling a 2-antilocal interleaved language. \textbf{Top:} three rotation classes (see Fig. \ref{fig:rotation_classes}) act as generators, each yielding a sequence of $m=2$ randomly sampled trigrams from within its class. \textbf{Middle:} a control sentence formed by concatenating components. \textbf{Bottom:} a 2-back sentence formed by interleaving components.}
  \label{fig:kBack_sampling}
\end{figure}

As illustrated in Figure~\ref{fig:kBack_sampling}, to generate a $k$-\textit{back} language, we select $k+1$ rotation classes to act as trigram generators. Sentences of a $k$-back language are sampled in two steps. First, form $k+1$ base components (one for each generator) by concatenating \textit{m} trigrams randomly sampled from each generator. The base components are then interleaved. In the resulting sentences, $n$-grams are uniformly distributed for $n\leq k+1$ because any window size $n\leq k+1$ contains at most one symbol from any given generator, and unigrams are uniformly distributed within and across generators. We compare each of our $k$-back languages against a (non-$k$-antilocal) control language generated in the same way, except that instead of interleaving base components, we simply concatenate them.       

With this construction, we can generate languages at arbitrary $k$. In the experiments described below, we compare LLMs' performance learning languages at different values of $k$ against the corresponding control languages. The control languages match the corresponding $k$-antilocal languages for global entropy (because target-language sentences and control sentences are simply permutations of each other). Appendix~\ref{apx: kback generation} provides more details about language construction.

One motivation for the $k$-back construction is that these languages contain dependencies only of length $k+1$. This controls for a potential confound in earlier work based on natural language datasets, 
where lengths of dependencies are unevenly distributed. Models are thus trained on many more short-dependency exemplars than they are on longer-range dependencies, which could partly explain the apparent bias towards information locality. Here, differences in model performance cannot be attributed to skewed representation of dependency lengths in training data.
\section{Experiments}

\begin{figure*}
  \includegraphics[width=\textwidth]{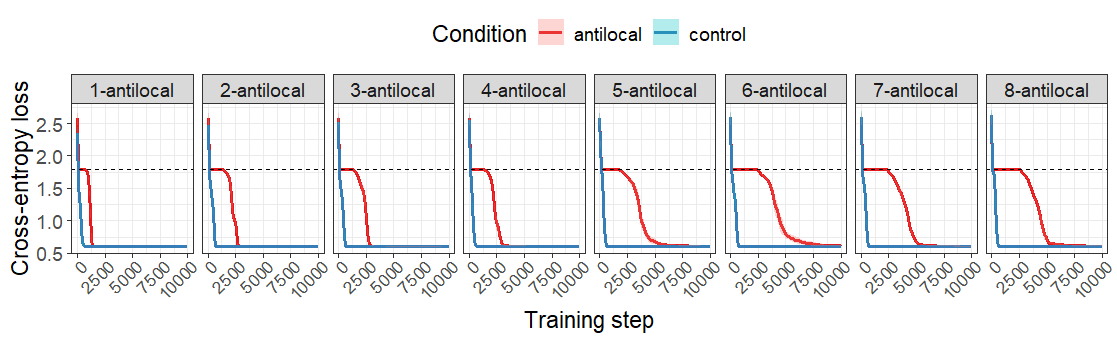}

  \vspace{-5pt}
  
  \caption{Cross-entropy loss on validation data from GPT-2 models trained on $k$-antilocal and matched control languages. We plot the first 10K training steps, after which values become too small for meaningful visualization. The dotted line marks $\ln{6}$ (which is the loss we expect from random guessing among vocabulary items).}
  \label{fig:loss plot}
\end{figure*}

We generate $k$-back languages for 8 values of $k$ from 1 through 8. For each language, we sample 100,000 sentences of approximately 1,000 tokens in length to yield a corpus of approximately 100 million tokens, reserving 80,000 for training, 10,000 for validation, and 10,000 for testing (unused).

We train randomly initialized transformer-based GPT-2 language models (small, 124M parameters; \citealp{radford2019language}) on these languages. Models use custom tokenizers with only the small vocabularies in our antilocal languages and required special tokens.
We repeat training across 3 random seeds, evaluating on validation data every 100 training steps.
Results for each evaluation step are reported as the average cross-entropy loss across seeds.
Appendix~\ref{apx:model training details} provides more training details.

\subsection{Learning Success}
As shown in Figure~\ref{fig:loss plot}, all models converged to approximately the same cross-entropy loss regardless of $k$ and whether being trained on a $k$-back or control language. Models' minimum losses consistently fell within 1\% of the minimum possible cross-entropy. This is specifically defined as the average per-token entropy of the $k$-back languages' true probability distribution (approximately 0.596 for all $k$, as derived in Appendix~\ref{apx: kback generation}). Altogether, these results demonstrate no significant effect of (anti-)locality on GPT-2's learning success, at least to the degree that our idealized datasets explore it.

\subsection{Learning Speed}
We also observe a marked difference in learning speed between $k$-back and paired control languages.
Figure~\ref{fig:difference_plot} plots their difference in validation cross-entropy loss through training. For all values of $k$, we see a significant period of divergence in losses during the first 5,000 training steps, where models trained on $k$-back languages have much larger loss than those trained on paired controls. Furthermore, as $k$ increases, this period also generally lengthens due to the longer time it takes for $k$-back forms of languages to converge (as observed in Figure~\ref{fig:loss plot}). This suggests that increased antilocality indeed makes language learning significantly more difficult for GPT-2 in terms of speed.
\section{Discussion}

Our results support the conclusion of prior work \cite{KalliniEtAl:2024, SomeyaEtAl:2025} that LLMs like GPT-2 have a learning bias toward information locality, i.e., languages with more local statistical structure are easier for these learners. In our results, this bias is reflected in learning speed. In all cases, models converged on control languages fairly quickly, with antilocal languages achieving comparable performance markedly more slowly. Further, the delay increased with $k$. These findings are consistent with the claim that more antilocal languages are more difficult for GPT-2 to learn.

Interestingly, unlike \citet{KalliniEtAl:2024} and \citet{SomeyaEtAl:2025}, we did not find an effect of locality on achievable cross-entropy loss (or, by proxy, KL divergence). Our models achieved approximately optimal cross-entropy loss (KL divergence of $\sim~$0) on all languages, regardless of $k$. It is possible that differences in achieved cross-entropy loss in prior work reflects differences other than information locality. One possibility is that the dependency-length distribution in prior training corpora is skewed towards shorter dependencies. Each corpus we tested here had a fixed dependency length throughout. Further work would be required to explore the extent to which this difference is relevant to explaining the relevant results. In any case, if it is true that information locality does not directly constrain achievable cross-entropy loss but does constrain learning speed (with more information-local languages being more quickly learnable), this raises questions for the interpretation of prior results where effects are observed \textit{both} in terms of learning speed \textit{and} in terms of minimum cross-entropy loss. Again, we leave such questions for future work.

 Finally, it remains unclear exactly what mechanism is responsible for the information-locality bias in GPT-2, but one possibility is that it depends on GPT's autoregressive (i.e., incremental) design. This idea is supported by results from DeBERTa-v3~\cite{he2021debertav3,he2021deberta}, a masked language model (see Appendix~\ref{apx:deberta results}), which we found to learn $k$-antilocal languages as easily as their matched controls (though there are other differences between GPT-2 and DeBERTa, such as the implementation of position embeddings). If so, it is promising for the prospect of comparative cognitive science involving humans and autoregressive LLMs \citep{FutrellMahowald:2026}: the functional constraints of autoregression apply to humans as well as LLMs, suggesting that conclusions about LLMs' antilocality bias could also be relevant for the analogous bias in humans. Nevertheless, much more work remains to be done before the merits of this emerging research program can be fully appreciated.
\section*{Limitations}

Our work has some limitations worth noting. First, we only explore one type of antilocality, achieved by interleaving trigrams from distinct generators. There are certainly other means to create antilocal languages, and training LLMs on them may yield additional insights.

Within our setting, it is possible that by increasing $k$ further, we could observe different effects.
Notably, exploring this would require generating considerably longer sequences, and thus it would become significantly more compute-intensive to train LLMs on them, which is why we selected our range of $k$ values as presented in this work.
There are other generation parameters whose effects could be explored as well, e.g., the base $n$-gram size (we only used trigrams), the vocabulary size (we only had six tokens), the size of generators (ours are single equivalence classes, but in principle they could be larger), and the length of sentences (our sentences are all $\sim$1,000 tokens long). 

One may also question the extent to which results based on languages as simple as the ones considered here have implications for LLMs' learning of human language, which is much more complex.
Additionally, our antilocal languages differ from their controls not only in terms of $k$-antilocality, but in other ways as well (e.g. $k$-back languages contain crossing dependencies, but the controls do not), which may or may not be relevant. Our results must therefore be interpreted with caution, with an eye to potential confounds.

Furthermore, this work only trains one GPT-2 and one DeBERTa architecture of comparable size (complexity). Training model variants at additional sizes may yield insights into how scaling model architectures affects locality biases. 
Similarly, this work is unable to answer fine-grained questions about how architectural properties (e.g., positional encoding type) influence LLMs' locality bias.

This work aimed to take the first step in understanding locality effects on LLMs' learning in a highly controlled and structured setting, thus we considered such inquiry out of scope.
Future work should consider exploring more variations on this setting to discover deeper insights into the nature of LLMs' apparent learning bias toward locality.



\bibliography{custom}

@book{Chomsky:2000,
    author  = {Noam Chomsky},
    title   = {New Horizons in the Study of Language and Mind},
    year    = "2000",
    publisher = {Cambridge University Press},
    address = {Cambridge}
}

@book{Chomsky:1965,
    author  = {Noam Chomsky},
    title   = {Aspects of the Theory of Syntax},
    year    = "1965",
    publisher = {MIT Press},
    address = {Cambridge, MA}
}

@book{Lenneberg:1967,
    author  = {Eric Lenneberg},
    title   = {Biological Foundations of Language},
    year    = "1967",
    publisher = {John Wiley and Sons},
    address = {New York}
}

@book{Moro:2016,
    author  = {Andrea Moro},
    title   = {Impossible Languages},
    year    = "2016",
    publisher = {MIT Press},
    address = {Cambridge, MA}
}

@article{Chomskyetal:2023,
	author = {Noam Chomsky and Ian Roberts and Jeffrey Watumull},
	year = "2023",
	title = {The False Promise of {ChatGPT}},
	journal = {The New York Times},
	url = "https://www.nytimes.com/2023/03/08/opinion/noam-chomsky-chatgpt-ai.html",
}

@article{SartoriOrru:2023,
	author = {Giuseppe Sartori and Graziella Orrù},
	year = "2023",
	title = {Language models and psychological sciences},
	journal = {Frontiers in Psychology},
    volume = "14",
    pages = "1279317"
}

@article{FriedericiEtAl:2017,
	author = {Angela D. Friederici and Noam Chomsky and Robert C. Berwick and Andrea Moro and Johan J. Bolhuis},
	year = "2017",
	title = {Language, mind and brain},
	journal = {Nature Human Behavior},
    volume = "1",
    pages = "713--722"
}

@article{MoroEtAl:2023,
	author = {Andrea Moro and Matteo Greco and Stefano F. Cappa},
	year = "2023",
	title = {Large languages, impossible languages and human brains},
	journal = {Cortex},
    volume = "167",
    pages = "82--85"
}

@article{FutrellMahowald:2026,
	author = {Richard Futrell and Kyle Mahowald},
	year = "2026",
	title = {How linguistics learned to stop worrying and love the language models},
	journal = {Behavioral and Brain Sciences},
    volume = "49",
    pages = "e198"
}

@article{BowersMitchell:2026,
	author = {Jeffrey S. Bowers and Jeff Mitchell},
	year = "2026",
	title = {Studies with impossible languages falsify {LMs} as models of human language},
	journal = {Behavioral and Brain Sciences},
    volume = "49",
    pages = "e202"
}

@article{Bowers:2026,
	author = {Jeffrey S. Bowers},
	year = "2026",
	title = {The successes and failures of artificial neural networks (ANNs) highlight the importance of innate linguistic priors for human language acquisition},
	journal = {Psychological Review},
    volume = "133",
    issue = "4",
    pages = "994--1005"
}

@article{XuEtAl:2026,
    author = {Tianyang Xu and Tatsuki Kuribayashi and Yohei Oseki and
Ryan Cotterell and Alex Warstadt},
    title = "Can Language Models Learn Typologically Implausible Languages?",
    journal = {Transactions of the Association for Computational Linguistics},
    year = "2026",
    volume = "14",
    pages = "588–611"
}

@inproceedings{ZivEtAl:2026,
    author = {Imry Ziv and Nur Lan and Emmanuel Chemla},
    title = "Biasless Language Models Learn Unnaturally: How LLMs Fail to
Distinguish the Possible from the Impossible",
    booktitle = "Proceedings of the 19th Conference of the European Chapter of the Association for Computational Linguistics
Volume 1: Long Papers",
    year = 2026,
    pages = "5393--5403"
}

@inproceedings{YangEtAl:2025,
    author = "Yang, Xiulin  and
      Aoyama, Tatsuya  and
      Yao, Yuekun  and
      Wilcox, Ethan Gotlieb",
    title = "Anything Goes? A Crosslinguistic Study of (Im)possible Language Learning in {LM}s",
    booktitle = "Proceedings of the 63rd Annual Meeting of the Association for Computational Linguistics (Volume 1: Long Papers)",
    year = 2025,
    pages = "26058--26077"
}

@inproceedings{devlin-etal-2019-bert,
    title = "{BERT}: Pre-training of Deep Bidirectional Transformers for Language Understanding",
    author = "Devlin, Jacob  and
      Chang, Ming-Wei  and
      Lee, Kenton  and
      Toutanova, Kristina",
    editor = "Burstein, Jill  and
      Doran, Christy  and
      Solorio, Thamar",
    booktitle = "Proceedings of the 2019 Conference of the North {A}merican Chapter of the Association for Computational Linguistics: Human Language Technologies, Volume 1 (Long and Short Papers)",
    month = jun,
    year = "2019",
    address = "Minneapolis, Minnesota",
    publisher = "Association for Computational Linguistics",
    url = "https://aclanthology.org/N19-1423/",
    doi = "10.18653/v1/N19-1423",
    pages = "4171--4186"
}

@article{DBLP:journals/corr/abs-1907-11692,
  author    = {Yinhan Liu and
               Myle Ott and
               Naman Goyal and
               Jingfei Du and
               Mandar Joshi and
               Danqi Chen and
               Omer Levy and
               Mike Lewis and
               Luke Zettlemoyer and
               Veselin Stoyanov},
  title     = {RoBERTa: {A} Robustly Optimized {BERT} Pretraining Approach},
  journal   = {CoRR},
  volume    = {abs/1907.11692},
  year      = {2019},
  url       = {http://arxiv.org/abs/1907.11692},
  archivePrefix = {arXiv},
  eprint    = {1907.11692},
  bibsource = {dblp computer science bibliography, https://dblp.org}
}

@inproceedings{wolf-etal-2020-transformers,
    title = "Transformers: State-of-the-Art Natural Language Processing",
    author = "Wolf, Thomas  and
      Debut, Lysandre  and
      Sanh, Victor  and
      Chaumond, Julien  and
      Delangue, Clement  and
      Moi, Anthony  and
      Cistac, Pierric  and
      Rault, Tim  and
      Louf, Remi  and
      Funtowicz, Morgan  and
      Davison, Joe  and
      Shleifer, Sam  and
      von Platen, Patrick  and
      Ma, Clara  and
      Jernite, Yacine  and
      Plu, Julien  and
      Xu, Canwen  and
      Le Scao, Teven  and
      Gugger, Sylvain  and
      Drame, Mariama  and
      Lhoest, Quentin  and
      Rush, Alexander",
    editor = "Liu, Qun  and
      Schlangen, David",
    booktitle = "Proceedings of the 2020 Conference on Empirical Methods in Natural Language Processing: System Demonstrations",
    month = oct,
    year = "2020",
    address = "Online",
    publisher = "Association for Computational Linguistics",
    url = "https://aclanthology.org/2020.emnlp-demos.6/",
    doi = "10.18653/v1/2020.emnlp-demos.6",
    pages = "38--45"
}

@misc{he2021debertav3,
      title={DeBERTaV3: Improving DeBERTa using ELECTRA-Style Pre-Training with Gradient-Disentangled Embedding Sharing}, 
      author={Pengcheng He and Jianfeng Gao and Weizhu Chen},
      year={2021},
      eprint={2111.09543},
      archivePrefix={arXiv},
      primaryClass={cs.CL}
}

@misc{radford2019language,
  title={Language Models are Unsupervised Multitask Learners},
  author={Radford, Alec and Wu, Jeff and Child, Rewon and Luan, David and Amodei, Dario and Sutskever, Ilya},
  year={2019},
}

@inproceedings{
he2021deberta,
title={DEBERTA: DECODING-ENHANCED BERT WITH DISENTANGLED ATTENTION},
author={Pengcheng He and Xiaodong Liu and Jianfeng Gao and Weizhu Chen},
booktitle={International Conference on Learning Representations},
year={2021},
url={https://openreview.net/forum?id=XPZIaotutsD}
}

@inproceedings{KalliniEtAl:2024,
    author = {Julie Kallini and Isabel Papadimitriou and Richard Futrell and Kyle Mahowald and Christopher Potts},
    title = "Mission: Impossible language models",
    booktitle = "Proceedings of the 62nd Annual Meeting of the Association for Computational Linguistics (Volume 1: Long Papers)",
    year = 2024,
    pages = "14691--14714"
}

@inproceedings{SomeyaEtAl:2025,
    author = {Taiga Someya and Anej Svete and Brian DuSell and Timothy J. O'Donnell and Mario Giulianelli and Ryan Cotterell},
    title = "Information Locality as an Inductive Bias for Neural Language Models",
    booktitle = "Proceedings of the 63rd Annual Meeting of the Association for Computational Linguistics (Volume 1: Long Papers)",
    year = 2025,
    pages = "27995--28013"
}

@incollection{Piantadosi:2024,
	author = {Stephen Piantadosi},
	year = "2024",
	title = {Modern language models refute Chomsky's approach to language},
	booktitle = {From fieldwork to linguistic theory: A tribute to Dan Everett},
    editors = {Edward Gibson and Moshe Poliak},
    publisher = "Language Science Press",
    address = "Berlin",
    pages = "353--414"
}

@phdthesis{Futrell:2017,
    author = "Richard Futrell",
    title = "Memory and locality in natural language",
    school = "MIT",
    year = "2017"
}

\appendix

\section{Antilocality and $N$-gram Uniformity}\label{apx:antilocality}

Given random variables $W$ and $C_n$ (over symbols drawn from vocabulary $V$ and over $n$-length context windows for $W$ respectively), we say that language $\mathcal{L}$ on $V^*$ is $k$-antilocal iff condition \eqref{k_antilocality_def} holds.

\begin{equation}\label{k_antilocality_def}
    I(W; C_n) = 0 \text{ for any } n\leq k
\end{equation}

\noindent In section \ref{sec:antilocal_languages}, we claimed that the languages we constructed for this paper are $k$-antilocal because their $n$-gram distributions are uniform for all $n\leq k+1$. Information theorists may find this trivial, but we demonstrate the required implication here for our own sakes.

First, by the definition of mutual information,

\begin{align}
    I(W; C_n) &= 0 \\
    &\Updownarrow \nonumber\\
    H(W,C_n) &= H(W) + H(C_n)\label{consequent}
\end{align}

\noindent Also, if $(n+1)$-gram distribution is uniform, then condition \eqref{antecedent} holds.

\begin{align}\label{antecedent}
    p(W=w|C_n=c_n)=p(W=w)&\\
    \text{ for all }&w, c_n \nonumber
\end{align}

\noindent Therefore, we need to show that \eqref{antecedent} implies \eqref{consequent}. This can be shown algebraically. First observe (implicitly quantified for all $w$ and for all $c_n$):

\begin{align}
    p(w|c_n) &= p(w)\\
    &\Updownarrow \nonumber\\
    \log p(w|c_n) &= \log p(w)\\
    &\Updownarrow \nonumber\\
    \log p(w,c_n) - &\log p(c_n) = \log p(w)\\
    &\Updownarrow \nonumber\\
    \log p(w,c_n) = &\log p(w) + \log p(c_n)\label{lemma1}
\end{align}

\noindent If equation \eqref{lemma1} holds for all $w, c_n$, then the expectation over both sides is also equal, so we have:

\begin{align}
    &\sum_{w,c_n} p(w,c_n) \log p(w,c_n)\\
    &= \sum_w p(w) \log p(w) + \sum_{c_n} p(c_n)\log p(c_n)\nonumber
\end{align}

\noindent Negating both sides gives:

\begin{equation}
    H(W,C_n) = H(W) + H(C_n)
\end{equation}

\noindent which is what we wanted to show.

\section{Language Generation}
\label{sec:kback_appendix}\label{apx: kback generation}
We generated $k$-back languages for 8 values of $k$. For each language, we sampled 100,000 sentences of approximately 1,000 tokens in length to yield a corpus of approximately 100 million tokens. For each $k$, we first select $k+1$ generators from the set of rotation classes on vocabulary $V=\{a,b,c,d,e,f\}$. Then, we sampled 100,000 sentences as follows. For each of the generators, randomly sample $m$ of its trigrams, and concatenate them. This yields $k+1$ base components of length $3m$ each. The base components are used to form a $k$-back sentence by interleaving them (i.e. taking the first symbol of each base component, then the second symbol of each base component, etc.). The same base components are used to form a control sentence by simply concatenating them. The sampling procedure for a $k$-back sentence (and the corresponding control) is illustrated graphically in Figure \ref{fig:kBack_sampling}.


For each value of $k$, we calculated a value of $m$ that would yield sentences of length $L_{\text{sentence}}=1,000$.

\begin{equation}\label{kback_m}
    m = \lfloor\frac{L_{\text{sentence}}-2}{3(k+1)}\rfloor
\end{equation} 

\noindent This follows from the formula for sentence length given below, where the added factor of 2 comes from the addition of start and end symbols to each sentence.

\begin{equation}\label{kback_length}
    L_{\text{sentence}} = 3m(k+1) + 2
\end{equation} 

\noindent We take the floor in \eqref{kback_m} because we need $m$ to be a whole number. The features of each of our $k$-back languages are given in Table \ref{tab:kback}.

\begin{table}
  \centering
  \begin{tabular}{cccc}
    \hline
    \textbf{k} & \textbf{m} & \vtop{\hbox{\strut \textbf{sentence}}\hbox{\strut \textbf{length}}} & \vtop{\hbox{\strut \textbf{total}}\hbox{\strut \textbf{tokens}}}\\
    \hline
    1     & 166 & 998 & $9.98\times10^7$           \\
    2     & 110 & 992 & $9.92\times10^7$         \\
    3     & 83 & 998  & $9.98\times10^7$        \\
    4     & 66 & 992  & $9.92\times10^7$        \\
    5      & 55 & 992  & $9.92\times10^7$         \\
    6     & 47  & 989  & $9.89\times10^7$       \\
    7     & 41  & 986   & $9.86\times10^7$      \\
    8     & 36 & 974  & $9.74\times10^7$        \\\hline
  \end{tabular}
  \caption{Features of our constructed $k$-back languages. All languages used trigrams ($n=3$) on the vocabulary $V=\{a,b,c,d,e,f\}$.}
  \label{tab:kback}
\end{table}

Note that our $k$-back languages all have approximately equal entropy. All sentences in a $k$-back language are equiprobable, so the entropy of a $k$-back language $\mathcal{L}$ is $H(\mathcal{L})=\log |\mathcal{L}|$. The number of sentences in a $k$-back language $\mathcal{L}$ formed on vocabulary $V$ is given by:

\begin{equation}
    |\mathcal{L}| = |V|^{m(k+1)}
\end{equation}

\noindent So entropy is $H(\mathcal{L})=m(k+1) \log |V|$. However, equation \eqref{kback_m} shows how $m$ can be approximately expressed (ignoring floor) as a multiple of $\frac{1}{k+1}$. The entropy is therefore roughly constant across languages as:

\begin{equation}
    H(\mathcal{L})\approx\frac{L_{\text{sentence}}-2}{3}\log |V|
\end{equation}

\noindent Note that we report average \textit{per-token} cross-entropy in our results. Ideally, that is simply the quotient $H(\mathcal{L})/1000$ ($\approx 0.596$). However, sentence length actually varies slightly across values of $k$ (this is because of the floor operator in our calculation of $m$; see equation \eqref{kback_m}), so the true per-token entropy is slightly unstable.

Crucially for our purposes, a $k$-back language has uniform $n$-gram distributions for $n\leq k+1$. To see this, first note that the transitions between trigrams in the control language (simple concatenation of trigrams) are entirely random: because generators are based on rotation classes, the first symbol of the next trigram can be any vocabulary item with equal probability. In a $k$-back language, all these transitions adjacent (the first symbols of the first $k+1$ trigrams become adjacent), which yields a sequence of $k+1$ independent uniformly distributed vocabulary items. Shifting the selected window of $k+1$ symbols to the right by $i$ positions removes symbols from the leftmost $i$ generators in the sequence only to add them back on the right. Any window of $k+1$ or fewer symbols (not counting start and end tokens) is thus effectively a random sequence of uniformly distributed vocabulary items. Non-uniformity is possible only with windows of at least $k+2$ symbols, because here the same generator can be represented more than once.


\section{Model Training Details}\label{apx:model training details}

Each GPT-2 model was trained using Hugging Face Trainer\footnote{\url{https://huggingface.co/docs/transformers/en/main_classes/trainer}} on a single NVIDIA A40 GPU with a batch size of 16 (maximized for available VRAM), maximum learning rate of $1 \times 10^{-5}$ and a configured maximum sequence length of 1,024 (sufficient for the sequences in our language).
Similar to prior work in this area \cite{KalliniEtAl:2024,YangEtAl:2025}, we apply a constant learning rate schedule with linear warmup for the first 1,000 steps.\footnote{While these prior works applied a 10\% warmup phase, we found that this significantly increaed training time without improving performance on our languages (which models generally achieved near optimal cross-entropy on regardless).}
Based on preliminary experiments, models are trained for a maximum of 4 epochs, which was found sufficient for convergence on the hardest $k$-back language in preliminary tests.
For both models, gradient norms are clipped to 1.0 for training stability.

In the reported results, models are evaluated on the held-out validation partition (10,000 examples) once every 100 training steps (batches), yielding 200 total evaluations through training.
For GPT-2, training stops early if the average evaluation cross-entropy loss is within 0.002 of the
analytical per-token entropy of the language\footnote{This entropy is approximately 0.596 across all values of $k$, derived in Appendix~\ref{apx: kback generation}.} for 4 consecutive evaluations. Where training stops early, subsequent losses are held constant at the model's last recorded loss. To speed up matrix multiplications, we use TensorFloat-32 (TF32) precision during training. All other training arguments have default values (for \texttt{transformers} 4.35.2; \citealp{wolf-etal-2020-transformers}).

\section{Supplementary DeBERTa Results}\label{apx:deberta results}

We report additional results on DeBERTa-v3~\cite{he2021debertav3,he2021deberta} (base, 184M parameters), a masked transformer-based language model. As GPT-2 is an autoregressive decoder-only model, this allows us to examine how the autoregressive property of GPT-2 impacts its bias toward locality observed in prior works. We use DeBERTa-v3's default mask probability of 15\% for training and evaluation. Masked models have lower expected entropy due to their access to forward and backward context, so we instead apply early stopping when the model's validation loss is within 0.0150 of the optimal loss for a model where 15\% of tokens are masked (0.0403).\footnote{As a masked token is only unpredictable if the two other tokens within the same trigram are simultaneously masked, we calculate the optimal loss as $0.15^2 \ln{6} \approx 0.0403$. The range of 0.0150 is set based on preliminary experiments without early stopping, where models generally appeared to converge within this range of losses. }
Otherwise, we use the same training configuration as described in Appendix~\ref{apx:model training details}.

\begin{figure*}
   \includegraphics[width=\textwidth]{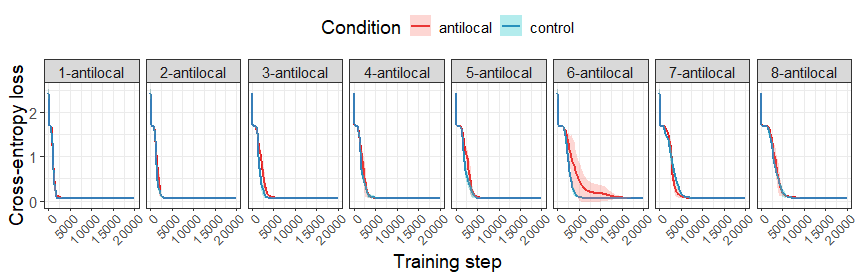}

  \vspace{-5pt}
  
  \caption{Cross-entropy loss on validation data from DeBERTa-v3 models trained on $k$-antilocal and matched control languages. Results are averaged across three seeds, with error ribbons showing standard deviation across seeds.}
  \label{fig:deberta loss}
\end{figure*}

\begin{figure*}
   \includegraphics[width=\textwidth]{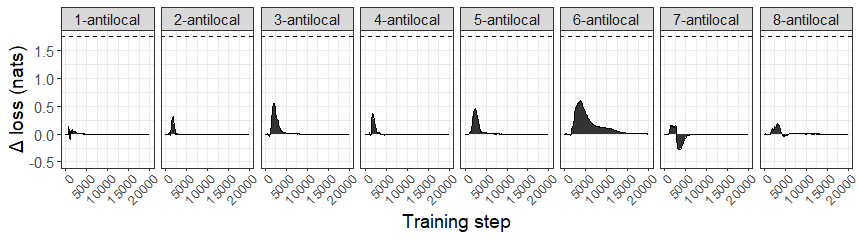}

  \vspace{-5pt}
  
  \caption{Difference in DeBERTa-v3 validation cross-entropy loss while training on antilocal vs. control (i.e., simple trigram) languages (antilocal $-$ control). All models eventually achieve optimal performance on all languages. The dotted line marks maximum expected difference, in which the control models perform optimally while the antilocal models perform at chance.}
  \label{fig:deberta diff}
\end{figure*}

\paragraph{Results.}
Comparable results for DeBERTa-v3 are presented in Figures~\ref{fig:deberta loss} and \ref{fig:deberta diff}. Like GPT-2, we observe that DeBERTa-v3 achieves comparable cross-entropy loss across all languages regardless of antilocality. Also like GPT-2, languages are learned slightly more slowly with higher values of $k$, though this slowdown occurs with both $k$-back languages and their corresponding controls, and hence cannot be attributed to the degree of antilocality. The global slowdown with $k$ is presumably due to the fact that higher values of $k$ require more distinct generators (both for $k$-back languages and the controls). Higher values of $k$ thus introduce more distinct patterns for models to learn, making the slowdown unsurprising. 

Unlike with GPT-2, DeBERTa learns a $k$-back language about as quickly as the corresponding (non-antilocal) control language. Although in some instances (e.g., $k=6$), our models learned control languages slightly faster on average, there is no clear trend for more antilocal languages to be learned more slowly. Additionally, with GPT-2 we observed a contrast in learning plateaus: there were extended periods in training where models exhibited loss consistent with random guessing across vocabulary items, but these plateaus occurred only with $k$-back languages, and not with the controls. With DeBERTa, we do not observe any distinction like this. We take these results to suggest that the bias for information locality, if it exists at all in DeBERTa, is quite weak.

\paragraph{Model choice.}
Lastly, we note here that DeBERTa-v3 was intentionally chosen over earlier masked LMs such as BERT~\cite{devlin-etal-2019-bert} or RoBERTa~\cite{DBLP:journals/corr/abs-1907-11692}, which did not converge on our languages.\footnote{More specifically, they only reached a minimum cross-entropy loss equivalent to that of a uniform distribution over the six main content tokens in our languages.} We suspect that this is due to the random initialization of embeddings, which causes low cosine similarity between them and the expected values of attention weights at the first layer to be a uniform distribution. Within the first layer, each token's embedding would thus be computed as a perfect mix of every other token's embedding (all of which are calculated by adding positional and content embeddings). Since each sequence contains the entire vocabulary of the dataset with nearly equal frequency, this destroys all meaningful contextual information in the sequences from the very beginning of the architecture. 

GPT-2 avoids this issue by being autoregressive, thus being exposed to incrementally larger subsequences of each training sequence in parallel and enabling learning of meaningful positional and content embeddings for each token. DeBERTa-v3 avoids this by maintaining separate positional and content embeddings throughout transformer layers, which are used for various attention operations. Notably, beyond its masked encoder-only training paradigm, this is another important structural difference of DeBERTa-v3 from GPT-2, thus the differences in results could be attributed in part to this.

\end{document}